\documentclass[10pt,twocolumn,letterpaper]{article}

\usepackage{cvpr}
\usepackage{times}
\usepackage{epsfig}
\usepackage{graphicx}
\usepackage{amsmath}
\usepackage{amssymb}

\usepackage{multirow}

\usepackage{algorithm}
\usepackage{algpseudocode}% http://ctan.org/pkg/algorithmicx
\algrenewcommand\algorithmicindent{.9em}%

\cvprfinalcopy % *** Uncomment this line for the final submission

\def\cvprPaperID{****} % *** Enter the CVPR Paper ID here

\newcommand{\tabincell}[2]{\begin{tabular}{@{}#1@{}}#2\end{tabular}}

\newlength\savewidth

\newcommand\shline{\noalign{\global\savewidth\arrayrulewidth
  \global\arrayrulewidth 1pt}\hline\noalign{\global\arrayrulewidth\savewidth}}

\makeatletter
\renewcommand\paragraph{\@startsection{paragraph}{4}{\z@}%
  {.5em \@plus1ex \@minus.2ex}%
  {-1em}%
  {\normalfont\normalsize\bfseries}}

\usepackage[pagebackref=true,breaklinks=true,letterpaper=true,colorlinks,bookmarks=false]{hyperref}

\ifcvprfinal\fi
\begin{document}

%%%%%%%%% TITLE
\title{Relation Networks for Object Detection}

\author{Han Hu$^1$\thanks{Equal contribution. \dag This work is done when Jiayuan Gu is an intern at Microsoft Research Asia.}\quad Jiayuan Gu$^{2*\dag}$ \quad Zheng Zhang$^{1*}$ \quad Jifeng Dai$^1$ \quad Yichen Wei$^1$ \vspace{8pt}\\
        $^1$Microsoft Research Asia\\
    $^2$Department of Machine Intelligence, School of EECS, Peking University\\
        {\tt\small \{hanhu,v-jiaygu,zhez,jifdai,yichenw\}@microsoft.com} \\
}

\maketitle

\begin{abstract}
Although it is well believed for years that modeling relations between objects would help object recognition,  there has not been evidence that the idea is working in the deep learning era. All state-of-the-art object detection systems still rely on recognizing object instances \textbf{individually}, without exploiting their relations during learning.

This work proposes an object relation module. It processes a set of objects \textbf{simultaneously} through interaction between their appearance feature and geometry, thus allowing modeling of their relations. It is lightweight and in-place. It does not require additional supervision and is easy to embed in existing networks. It is shown effective on improving object recognition and duplicate removal steps in the modern object detection pipeline. It verifies the efficacy of modeling object relations in CNN based detection. It gives rise to the \textbf{first fully end-to-end object detector}. Code is available at \url{https://github.com/msracver/Relation-Networks-for-Object-Detection}.

\vspace{-1em}
\end{abstract}

\vspace{-.5em}
\section{Introduction}
\label{sec.introduction}

Recent years have witnessed significant progress in object detection using deep convolutional neutral networks (CNNs)~\cite{huang2016speed}. The state-of-the-art object detection methods~\cite{he2014spatial,girshick2015fast,ren2015faster,dai2016rfcn,lin2016feature,dai2017deformable,he2017mask} mostly follow the \emph{region based} paradigm since it is established in the seminal work R-CNN~\cite{girshick2014rich}. Given a sparse set of region proposals, object classification and bounding box regression are performed on each proposal \emph{individually}. A \emph{heuristic and hand crafted} post-processing step, non-maximum suppression (NMS), is then applied to remove duplicate detections.

It has been well recognized in the vision community for years that contextual information, or \emph{relation} between objects, helps object recognition~\cite{divvala2009empirical,galleguillos2008object,torralba2003context,tu2008auto,shotton2006textonboost,mottaghi2014role,galleguillos2008object,galleguillos2010context,chen2017spatial}. Most such works are before the prevalence of deep learning. During the deep learning era, there is no significant progress about exploiting object relation for detection learning. Most methods still focus on recognizing objects separately.

One reason is that object-object relation is hard to model. The objects are at arbitrary image locations, of different scales, within different categories, and their number may vary across different images. The modern CNN based methods mostly have a simple regular network structure~\cite{he2016deep,he2017mask}. It is unclear how to accommodate above irregularities in existing methods.

Our approach is motivated by the success of attention modules in natural language processing field~\cite{britz2017massive,vaswani2017attention}. An attention module can effect an individual element (\eg, a word in the target sentence in machine translation) by aggregating information (or features) from a set of elements (\eg, all words in the source sentence). The aggregation weights are automatically learnt, driven by the task goal. An attention module can  model dependency between the elements, without making excessive assumptions on their locations and feature distributions. Recently, attention modules have been successfully applied in vision problems such as image captioning~\cite{xu2015show}.

In this work, for the first time we propose an adapted attention module for object detection. It is built upon a basic attention module. An apparent distinction is that the primitive elements are objects instead of words. The objects have 2D spatial arrangement and variations in scale/aspect ratio. Their locations, or geometric features in a general sense, play a more complex and important role than the word location in an 1D sentence. Accordingly, the proposed module extends the original attention weight into two components: the original weight and a new geometric weight. The latter models the spatial relationships between objects and only considers the \emph{relative geometry} between them, making the module \emph{translation invariant}, a desirable property for object recognition. The new geometric weight proves important in our experiments.

The module is called \emph{object relation module}. It shares the same advantages of an attention module. It takes variable number of inputs, runs in parallel (as opposed to sequential relation modeling~\cite{li2016attentive,stewart2016end,chen2017spatial}), is fully differentiable and is in-place (no dimension change between input and output). As a result, it serves as a basic building block that is usable in any architecture flexibly.

\setlength{\tabcolsep}{2pt}
\begin{figure}
\begin{center}
\begin{tabular}{c}
{\includegraphics[width=0.46\textwidth]{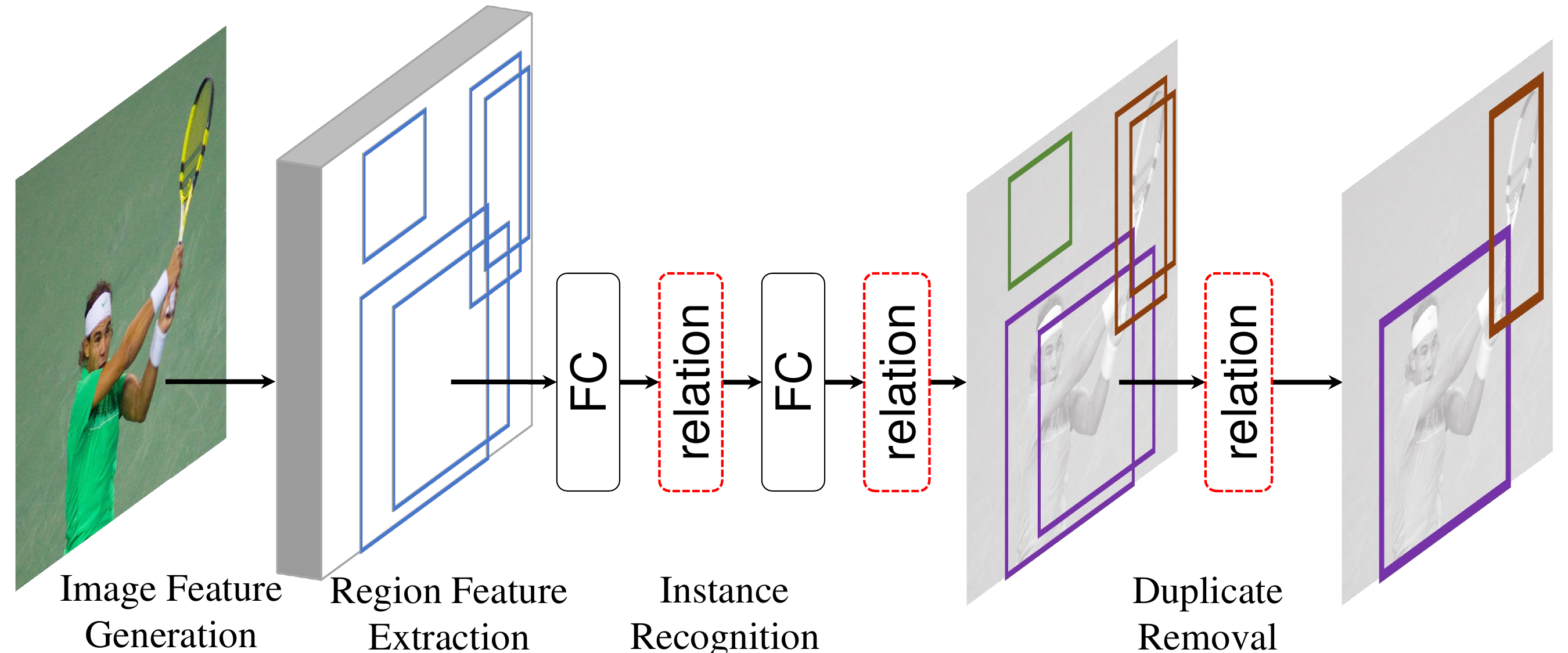}} \\
\end{tabular}
\end{center}
\vspace{-.5em}
\caption{Current state-of-the-art object detectors are based on a four-step pipeline. Our object relation module (illustrated as red dashed boxes) can be conveniently adopted to improve both instance recognition and duplicate removal steps, \emph{resulting in an end-to-end object detector}.}
\label{figure.introduction}
\vspace{-1em}
\end{figure}

Specifically, it is applied to several state-of-the-art object detection architectures~\cite{ren2015faster,dai2017deformable,lin2016feature} and show consistent improvement. As illustrated in Figure~\ref{figure.introduction}, it is applied to improve the \emph{instance recognition} step and learn the \emph{duplicate removal} step (see Section~\ref{sec.detection_pipeline_review} for details). For instance recognition, the relation module enables joint reasoning of all objects and improves recognition accuracy (Section~\ref{sec.feature_enhance}). For duplicate removal, the traditional NMS method is replaced and improved by a lightweight relation network (Section~\ref{sec.duplicate_removal}), resulting in \emph{the first end-to-end object detector} (Section~\ref{sec.end_to_end_detection}), to our best knowledge.

In principle, our approach is fundamentally different from and would complement most (if not all) CNN based object detection methods. It exploits a new dimension: \emph{a set of objects are processed, reasoned and affect each other simultaneously, instead of recognized individually}.

The object relation module is general and not limited to object detection. We do not see any reason preventing it from finding broader applications in vision tasks, such as instance segmentation~\cite{li2016fully}, action recognition~\cite{simonyan2014two}, object relationship detection~\cite{krishna2017visual}, caption~\cite{xu2015show}, VQA~\cite{antol2015vqa}, etc. Code is available at \url{https://github.com/msracver/Relation-Networks-for-Object-Detection}.

\vspace{-.5em}
\section{Related Works}

\vspace{-.5em}

\textbf{Object relation in post-processing} Most early works use object relations as a post-processing step~\cite{divvala2009empirical,galleguillos2008object,torralba2003context,tu2008auto,mottaghi2014role,galleguillos2008object}. The detected objects are re-scored by considering object relationships. For example, \textit{co-occurrence}, which indicates how likely two object classes can exist in a same image, is used by DPM~\cite{pedro2010dpm} to refine object scores. The subsequent approaches~\cite{Choi12tree,mottaghi2014role} try more complex relation models, by taking additional \emph{position} and \emph{size}~\cite{biederman1982scene} into account. We refer readers to \cite{galleguillos2010context} for a more detailed survey. These methods achieve moderate success in pre-deep learning era but do not prove effective in deep ConvNets. A possible reason is that deep ConvNets have implicitly incorporated contextual information by the large receptive field.

\textbf{Sequential relation modeling} Several recent works perform sequential reasoning (LSTM~\cite{li2016attentive,stewart2016end} and spatial memory network (SMN)~\cite{chen2017spatial}) to model object relations. During detection, objects detected earlier are used to help finding objects next. Training in such methods is usually sophisticated. More importantly, they do not show evidence of improving the state-of-the-art object detection approaches, which are simple feed-forward networks.

In contrast, our approach is parallel for multiple objects. It naturally fits into and improves modern object detectors.

\textbf{Human centered scenarios} Quite a few works focus on \emph{human-object} relation~\cite{yao12recognizing,GuptaHM15a,gkioxari2015contextual, DBLP:journals/corr/GkioxariGDH17}. They usually require additional annotations of relation, such as human action. In contrast, our approach is general for object-object relation and does not need additional supervision.

\textbf{Duplicate removal} In spite of the significant progress of object detection using deep learning, the most effective method for this task is still the greedy and hand-crafted non-maximum suppression (NMS) and its soft version~\cite{bodla2017soft}. This task naturally needs relation modeling. For example, NMS uses simple relations between bounding boxes and scores.

Recently, GossipNet~\cite{hosang2017learning} attempts to learn \emph{duplicate removal} by processing a set of objects as a whole, therefore sharing the similar spirit of ours. However, its network is specifically designed for the task and very complex (depth$>$80). Its accuracy is comparable to NMS but computation cost is demanding. Although it allows end-to-end learning in principle, no experimental evidence is shown.

In contrast, our relation module is simple, general and applied to duplicate removal as an application. Our network for duplicate removal is much simpler, has small computation overhead and surpasses SoftNMS~\cite{bodla2017soft}. More importantly, we show that an end-to-end object detection learning is feasible and effective, \emph{for the first time}.

\textbf{Attention modules in NLP and physical system modeling} Attention modules have recently been successfully applied in the NLP field~\cite{gehring2016convolutional,gehring2017convolutional,britz2017massive,vaswani2017attention} and in physical system modeling~\cite{battaglia2016interaction,watters2017visual,hoshen2017vain,santoro2017simple,DenilCCSF17,raposo2017discovering}. The attention module can well capture the long-term dependencies in these problems. In NLP, there is a recent trend of replacing recurrent neural networks by attention models, enabling parallelized implementations and more efficient learning~\cite{gehring2017convolutional,vaswani2017attention,DenilCCSF17}.

Our method is motivated by these works. We extend attention modeling to the important problem of object detection. For modeling visual object relations, their locations, or geometric features in a general sense,
play a  complex and important role. Accordingly, the proposed module introduces a novel geometric weight to capture the spatial relationship between objects. The novel geometric weight is translational invariant, which is an important property for visual modeling.

\begin{algorithm}[t]
\caption{Object relation module. Input is $N$ objects $\{(f_A^n, f_G^n)\}^N_{n=1}$. Dimension of  appearance feature $f_A$ is $d_f$. After each algorithm line is the computation complexity.}
\small
\begin{algorithmic}[1] % [1] for line numbers
\State \textbf{hyper param}: number of relations $N_r$
\State \textbf{hyper param} $d_k$: key feature dimension
\State \textbf{hyper param} $d_g$: geometric feature embedding dimension
\State \textbf{learnt weights}: $\{W_K^r,W_Q^r,W_V^r, W_G^r\}^{N_r}_{r=1}$

\For{every $(n,r)$} {\Comment{\scriptsize $O(NN_r)$}}
%\For{every $m$} %\Comment{$O(N)$}
\State compute $\{\omega_G^{mn,r}\}_{m=1}^N$ using Eq.~(\ref{eq.geometric_weight}) {\Comment{\scriptsize $O(N d_g)$}}
\State compute $\{\omega_A^{mn,r}\}_{m=1}^N$ using Eq.~(\ref{eq.appearance_weight}) {\Comment{\scriptsize $O(d_k(2d_f+N))$}}
%\EndFor

\State compute $\{\omega^{mn,r}\}_{m=1}^N$ using Eq.~(\ref{eq.object_relation_weight}) {\Comment{\scriptsize $O(N)$}}
\State compute $f_R^r(n)$ using Eq.~(\ref{eq.object_relation_module})  {\Comment{\scriptsize $O(d_f^2/N_r + Nd_f/N_r)$}}
\EndFor

\State \textbf{output} new feature $\{f_A^n\}^N_{n=1}$ using Eq.~(\ref{eq.object_relation_module_ensemble})

\end{algorithmic}
\label{alg.object_relation_module}
\end{algorithm}

\vspace{-.5em}
\section{Object Relation Module}
\label{sec.relation_module}
\vspace{-.5em}
We first review a basic attention module, called ``Scaled Dot-Product Attention''~\cite{vaswani2017attention}. The input consists of queries and keys of dimension $d_k$, and values of dimension $d_v$. Dot product is performed between the query and all keys to obtain their similarity. A softmax function is applied to obtain the weights on the values. Given a query $\mathbf{q}$, all keys (packed into matrices $K$) and values (packed into $V$), the output value is weighted average over input values,

\begin{equation}
v^{out} = softmax(\frac{\mathbf{q}K^t}{\sqrt{d_k}})V.
\label{eq.basic_attention}
\end{equation}

We now describe object relation computation. Let an object consists of its \emph{geometric} feature $\mathbf{f}_G$ and appearance feature $\mathbf{f}_A$. In this work, $\mathbf{f}_G$ is simply a 4-dimensional object bounding box and $\mathbf{f}_A$ is up to the task (Section~\ref{sec.feature_enhance} and ~\ref{sec.duplicate_removal}).

Given input set of $N$ objects $\{(\mathbf{f}_A^n, \mathbf{f}_G^n)\}^N_{n=1}$, the \emph{relation feature} $\mathbf{f}_R(n)$ of the whole object set with respect to the $n^{th}$ object, is computed as

\begin{equation}
\mathbf{f}_R(n) = \sum_{m} \omega^{mn}\cdot (W_V\cdot \mathbf{f}^m_{A}).
\label{eq.object_relation_module}
\end{equation}

The output is a weighted sum of appearance features from other objects, linearly transformed by $W_V$ (corresponding to values $V$ in Eq.~(\ref{eq.basic_attention})). The relation weight $\omega^{mn}$ indicates the impact from other objects. It is computed as

\begin{equation}
\omega^{mn} = \frac{\omega^{mn}_G \cdot \exp(\omega^{mn}_A)}{\sum_{k} \omega^{kn}_G \cdot \exp(\omega^{kn}_A)}.
\label{eq.object_relation_weight}
\end{equation}

\emph{Appearance weight} $\omega^{mn}_A$ is computed as dot product, similarly as in Eq.~(\ref{eq.basic_attention}),

\begin{equation}
\omega^{mn}_A = \frac{dot(W_K{\mathbf{f}}^{m}_A, W_Q \mathbf{f}^{n}_A)}{\sqrt{d_k}}.
\label{eq.appearance_weight}
\end{equation}

Both $W_K$ and $W_Q$ are matrices and play a similar role as $K$ and $Q$ in Eq.~(\ref{eq.basic_attention}). They project the original features $\mathbf{f}_A^m$ and $\mathbf{f}_A^n$ into subspaces to measure how well they match. The feature dimension after projection is $d_k$.

\emph{Geometry weight} is computed as

\begin{equation}
\omega^{mn}_G = \max \{0, W_G\cdot \mathcal{E}_G(\mathbf{f}_G^m, \mathbf{f}_G^n)\}.
\label{eq.geometric_weight}
\end{equation}

There are two steps. First, the geometry features of the two objects are embedded to a high-dimensional representation, denoted as $\mathcal{E}_G$. To make it invariant to translation and scale transformations, a 4-dimensional relative geometry feature is used, as $\left(\log(\frac{|x_m-x_n|}{w_m}), \log (\frac{|y_m-y_n|}{h_m}), \log(\frac{w_n}{w_m}), \log(\frac{h_n}{h_m})\right)^T$\footnote{It is a modified version of the widely used bounding box regression target~\cite{girshick2014rich}. The first two elements are transformed using $\log(\cdot)$ to count more on close-by objects. The intuition behind this modification is that we need to model distant objects while original bounding box regression only considers close-by objects.}. This 4-d feature is embedded to a high-dimensional representation by method in~\cite{vaswani2017attention}, which computes cosine and sine functions of different wavelengths. The feature dimension after embedding is $d_g$.

Second, the embedded feature is transformed by $W_G$ into a scalar weight and trimmed at 0, acting as a ReLU non-linearity. The zero trimming operation restricts relations only between objects of certain geometric relationships.

The usage of geometric weight Eq.~(\ref{eq.geometric_weight}) in the attention weight Eq.~(\ref{eq.object_relation_weight}) makes our approach distinct from the basic attention Eq.~(\ref{eq.basic_attention}). To validate the effectiveness of Eq.~(\ref{eq.geometric_weight}), we also experimented with two other simpler variants. The first is called \emph{none}. It does not use geometric weight Eq.~(\ref{eq.geometric_weight}). $\omega^{mn}_G$ is a constant 1.0 in Eq.~(\ref{eq.object_relation_weight}). The second is called \emph{unary}. It follows the recent approaches~\cite{duan2017one,vaswani2017attention}. Specifically, $\mathbf{f}_G$ is embedded into a high-dimension (same as $\mathbf{f}_A$) space in the same way~\cite{vaswani2017attention} and added onto $\mathbf{f}_A$ to form the new appearance feature. The attention weight is then computed as \emph{none} method. The effectiveness of our geometry weight is validated in Table~\ref{table.exp_feature_enhance_module_structure}(a) and Section~\ref{sec.exp_duplicate_removal}.

\setlength{\tabcolsep}{2pt}
\begin{figure}
\begin{center}
\begin{tabular}{c}
{\includegraphics[width=0.46\textwidth]{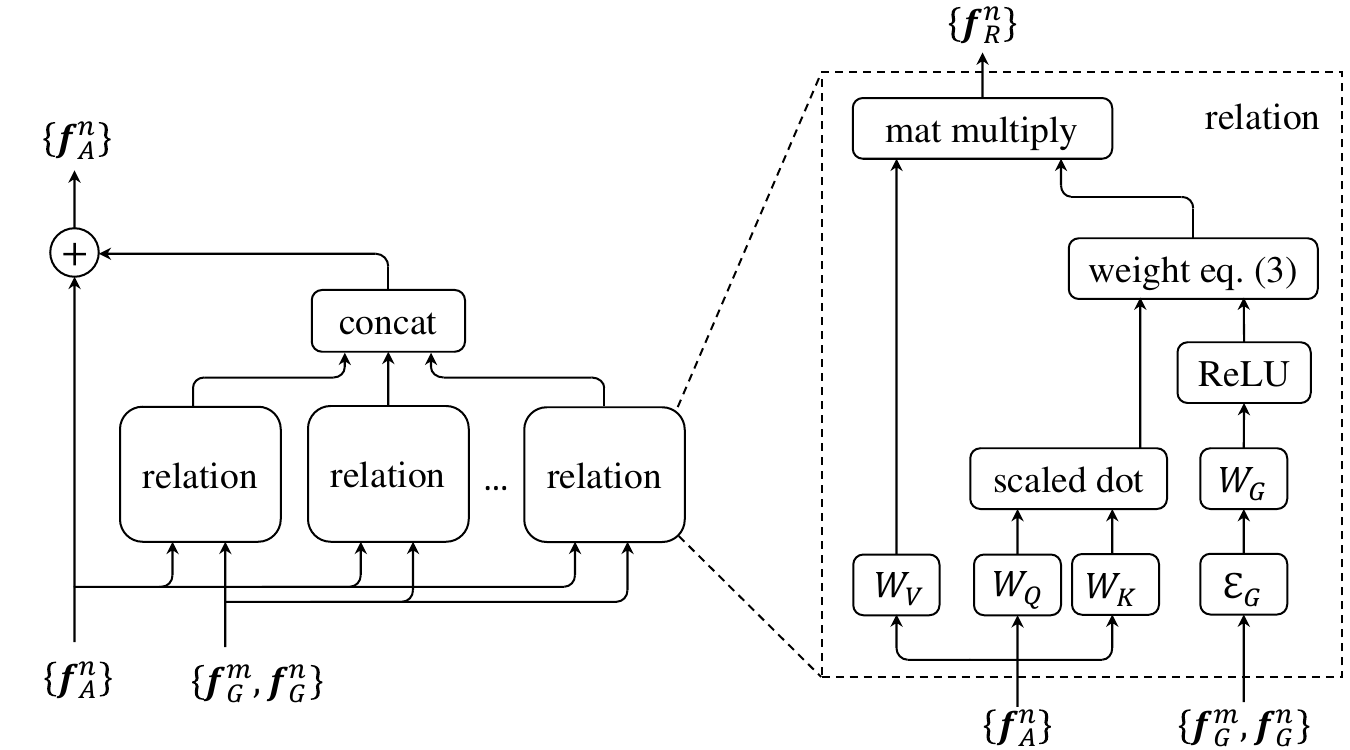}}\\
%{\includegraphics[width=0.26\textwidth]{figures/object_relation_module.pdf}} & {\includegraphics[width=0.2\textwidth]{figures/relation_compute_v3.pdf}} \\
\end{tabular}
\end{center}
\vspace{-.5em}
\caption{\textbf{Left}: object relation module as Eq.~(\ref{eq.object_relation_module_ensemble}); \textbf{Right}: relation feature computation as Eq.~(\ref{eq.object_relation_module}).}
\label{figure.object_relation_module}
\vspace{-1em}
\end{figure}

An \emph{object relation module} aggregates in total $N_r$ relation features and augments the input object's appearance feature via addition,

\begin{equation}
\mathbf{f}^{n}_A = \mathbf{f}^{n}_{A} + \text{Concat}[\mathbf{f}_R^1(n),...,\mathbf{f}_R^{N_r}(n)], \text{for all } n.
\label{eq.object_relation_module_ensemble}
\end{equation}

Concat($\cdot$) is used to aggregate multiple relation features\footnote{An alternative is Addition($\cdot$). However, its computation cost would be much higher because we have to match the channel dimensions of two terms in Eq.~(\ref{eq.object_relation_module_ensemble}). Only Concat($\cdot$) is experimented in this work.}. To match the channel dimension, the output channel of each $W_V^r$ is set as $\frac{1}{N_r}$ of the dimension of input feature $\mathbf{f}_A^m$.

The object relation module Eq.~(\ref{eq.object_relation_module_ensemble}) is summarized in Algorithm~\ref{alg.object_relation_module}. It is easy to implement using basic operators, as illustrated in Figure~\ref{figure.object_relation_module}.

Each relation function in Eq.~(\ref{eq.object_relation_module}) is parameterized by four matrices $(W_K,W_Q,W_G,W_V)$, in total $4N_r$. Let $d_f$ be the dimension of input feature $\mathbf{f}_A$.
The number of parameters is

\begin{equation}
O(\text{Space})=N_r(2d_f d_k+d_g)+d_f^2.
\label{eq.space_complexity}
\end{equation}

Following Algorithm~\ref{alg.object_relation_module}, the computation complexity is
\begin{equation}
O(\text{Comp.})=N d_f (2N_rd_k+d_f)+N^2N_r(d_g+d_k+d_f/N_r+1).
\label{eq.compute_complexity}
\end{equation}

Typical parameter value is $N_r=16$, $d_k=64$, $d_g=64$. In general, $N$ and $d_f$ are usually at the scale of hundreds. The overall computation overhead is low when applied to modern object detectors.

The relation module has the same input and output dimension, and hence can be regarded as a basic building block to be used in-place within any network architecture. It is fully differentiable, and thus can be easily optimized with back-propagation. Below it is applied in modern object detection systems.

\section{Relation Networks For Object Detection}
\label{sec.relation_for_detection}
\vspace{-.5em}

\subsection{Review of Object Detection Pipeline}
\label{sec.detection_pipeline_review}

This work conforms to the \emph{region based} object detection paradigm. The paradigm is established in the seminal work R-CNN~\cite{girshick2014rich} and includes majority of modern object detectors~\cite{he2014spatial,girshick2015fast,ren2015faster,dai2016rfcn,lin2016feature,dai2017deformable,he2017mask}\footnote{Another object detection paradigm is based on \emph{dense sliding windows}~\cite{liu2016ssd,redmon2016you,lin2017focal}. In this paradigm, the object number $N$ is much larger. Directly applying relation module as in this work is computationally costly. How to effectively model relations between dense objects is yet unclear.}. A four step pipeline is used in all previous works, as summarized here.

First step generates full image features. From the input image, a deep convolutional \emph{backbone} network extracts full resolution convolutional features (usually $16\times$ smaller than input image resolution). The backbone network~\cite{simonyan2015very,szegedy2015going,srivastava2015highway,he2016deep,chollet2016xception,zoph2017learning} is pre-trained on ImageNet classification task~\cite{deng2009imagenet} and fine-tuned during detection training.

Second step generates regional features. From the convolutional features and a sparse set of region proposals~\cite{uijlings2013selective,zitnick2014edge,ren2015faster}, a RoI pooling layer~\cite{he2014spatial,girshick2015fast,he2017mask} extracts fixed resolution regional features (\eg, $7\times7$) for each proposal.

Third step performs instance recognition. From each proposal's regional features, a \emph{head} network predicts the probabilities of the proposal belonging to certain object categories, and refine the proposal bounding box via regression. This network is usually shallow, randomly initialized, and jointly trained together with backbone network during detection training.

Last step performs duplicate removal. As each object should be detected only once, duplicated detections on the same object should be removed. This is usually implemented as a heuristic post-processing step called \emph{non-maximum suppression} (NMS). Although NMS works well in practice, it is manually designed and sub-optimal. It prohibits the end-to-end learning for object detection.

In this work, the proposed object relation module is used in the last two steps. We show that it enhances the \emph{instance recognition} (Section~\ref{sec.feature_enhance}) and learns \emph{duplicate removal} (Section~\ref{sec.duplicate_removal}). Both steps can be easily trained, either independently or jointly (Section~\ref{sec.end_to_end_detection}). The joint training further boosts the accuracy and gives rise to \emph{the first end-to-end general object detection system}.

\textbf{Our implementation of different architectures} To validate the effectiveness and generality of our approach, we experimented with different combination of state-of-the-art \emph{backbone} networks (ResNet~\cite{he2016deep}), and best-performing detection architectures including faster RCNN~\cite{ren2015faster}, feature pyramid networks (FPN)~\cite{lin2016feature}, and deformable convolutional network (DCN)~\cite{dai2017deformable}. Region proposal network (RPN)~\cite{ren2015faster} is used to generate proposals.

\begin{itemize}
\item \emph{Faster RCNN}~\cite{ren2015faster}. It is directly built on backbone networks such as ResNet~\cite{he2016deep}. Following~\cite{ren2015faster}, RPN is applied on the conv4 feature maps. Following~\cite{dai2017deformable}, the instance recognition head network is applied on a new 256-d $1\times1$ convolution layer added after conv5, for dimension reduction. Note that the stride in conv5 is changed from 2 to 1, as common practice~\cite{he2016deep}.

\item \emph{FPN}~\cite{lin2016feature}. Compared to Faster RCNN, it modifies the backbone network by adding top-down and lateral connections to build a feature pyramid that facilitates end-to-end learning across different scales. RPN and head networks are applied on features of all scales in the pyramid. We follow the training details in~\cite{lin2016feature}.

\item \emph{DCN}~\cite{dai2017deformable}. Compared to Faster RCNN, it modifies the backbone network by replacing the last few convolution layers in conv5 by deformable convolution layers. It also replace the standard RoI pooling by deformable RoI pooling. We follow the training details in~\cite{dai2017deformable}.
\end{itemize}

Despite the differences, a commonality in above architectures is that they all adopt the same head network structure, that is, the RoI pooled regional features undergo two fully connected layers (\emph{2fc}) to generate the final features for proposal classification and bounding box regression.

Below, we show that relation modules can enhance the instance recognition step using the \emph{2fc} head.

\subsection{Relation for Instance Recognition}
\label{sec.feature_enhance}

Given the RoI pooled features for $n^{th}$ proposal, two fc layers with dimension $1024$ are applied. The instance classification and bounding box regression are then performed via linear layers. This process is summarized as

\begin{equation}
\begin{tabular}{rll}
$RoI\_Feat_n$ & $\xrightarrow{FC}$ & $1024$ \\
       & $\xrightarrow{FC}$ &  $1024$ \\
       & $\xrightarrow{LINEAR}$ & ($score_n$, $bbox_n$)  \\
\end{tabular}
\label{eq.instance_recognition_2fc_head}
\end{equation}

\begin{figure}
\begin{center}
\begin{tabular}{cc}
{\includegraphics[width=0.16\textwidth]{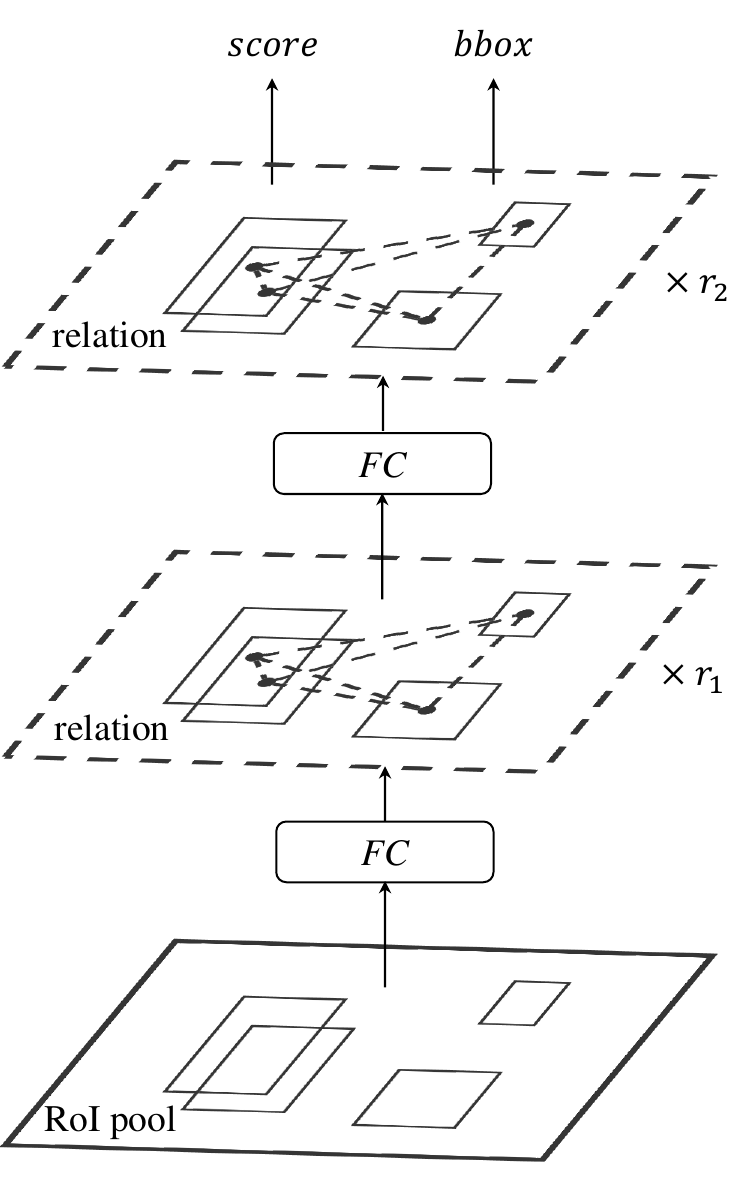}} & {\includegraphics[width=0.21\textwidth]{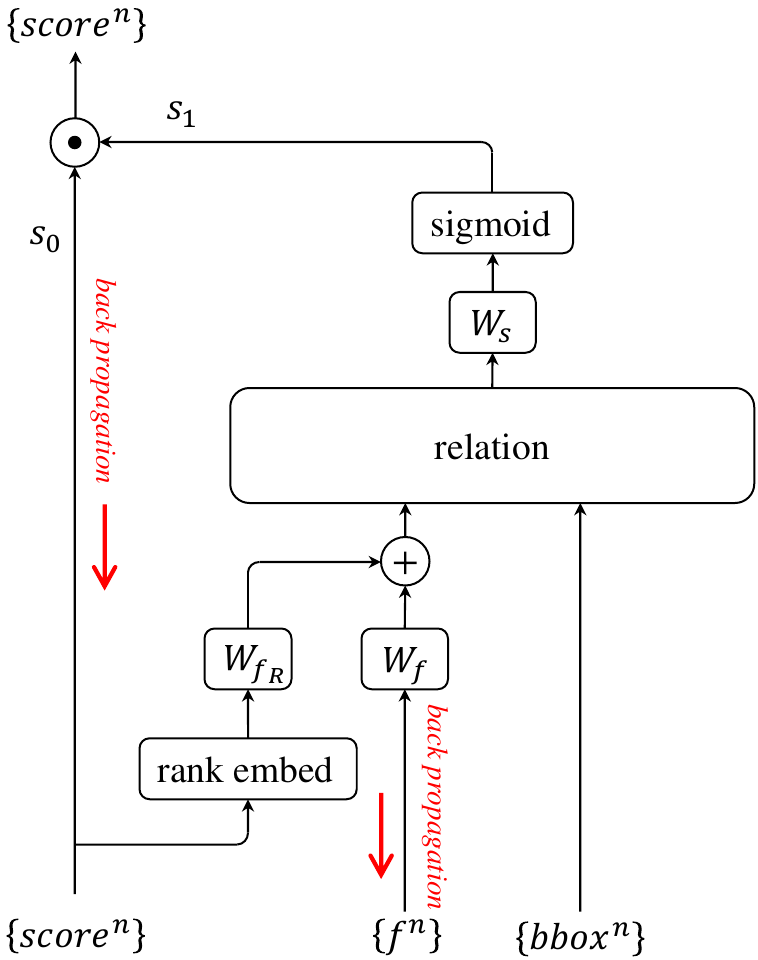}} \\
{(a) enhanced \emph{2fc} head} & {(b) duplicate removal network} \\
\end{tabular}
\end{center}
\caption{Illustration of enhanced \emph{2fc} head (a) and duplicate classification network (b) by object relation modules.}
\label{fig.duplicate_detection_with_relation}
\vspace{-1em}
\end{figure}

The object relation module (Section~\ref{sec.relation_module}, Algorithm~\ref{alg.object_relation_module}) can transform the $1024$-d features of all proposals without changing the feature dimension. Therefore, it can be used after either fc layer in Eq.~(\ref{eq.instance_recognition_2fc_head}) for arbitrary number of times\footnote{The relation module can also be used directly on the regional features. The high dimension ($256\times7^2=12544$ in our implementation), however, introduces large computational overhead. We did not do this experiment.}. Such enhanced \emph{2fc+RM} (RM for relation module) head is illustrated in Figure~\ref{fig.duplicate_detection_with_relation} (a) and summarized as

\begin{equation}
\begin{tabular}{rll}
$\{RoI\_Feat_n\}^N_{n=1}$ & $\xrightarrow{FC}$ & $1024 \cdot N$  $\xrightarrow{\{RM\}^{r_1}}$ $1024 \cdot N$ \\
       & $\xrightarrow{FC}$ &  $1024 \cdot N$ $\xrightarrow{\{RM\}^{r_2}}$ $1024 \cdot N$  \\
       & $\xrightarrow{LINEAR}$ & $\{(score_n, bbox_n)\}^N_{n=1}$  \\
\end{tabular}
\label{eq.instance_recognition_relation}
\end{equation}

In Eq.~(\ref{eq.instance_recognition_relation}), $r_{1}$ and $r_2$ indicate how many times a relation module is repeated. Note that a relation module also needs all proposals' bounding boxes as input. This notation is neglected here for clarify.

Adding relation modules can effectively enhance the instance recognition accuracy. This is verified via comprehensive ablation studies in experiments (Section~\ref{sec.exp_feature_enhance}).

\setlength{\tabcolsep}{0.5em}

\subsection{Relation for Duplicate Removal}
\label{sec.duplicate_removal}

The task of duplicate removal naturally requires exploiting the relation between objects. The heuristic NMS method is a simple example: the object with the highest score will erase its nearby objects (geometric relation) with inferior scores (score relation).

In spite of its simplicity, the greedy nature and manually chosen parameters in NMS makes it a clear sub-optimal choice. Below we show that the proposed relation module can learn to remove duplicate in a manner that is simple as well but more effective.

Duplicate removal is a two class classification problem. For each ground truth object, only one detected object matched to it is classified as \emph{correct}. Others matched to it are classified as \emph{duplicate}.

This classification is performed via a network, as illustrated in Figure~\ref{fig.duplicate_detection_with_relation} (b). The input is a set of detected objects (output from instance recognition, Eq.~(\ref{eq.instance_recognition_2fc_head}) or~(\ref{eq.instance_recognition_relation})). Each object has its final $1024$-d feature, the classification score $s_0$, and bounding box. The network outputs a binary classification probability $s_1$ $\in[0, 1]$ (1 for \emph{correct} and 0 for \emph{duplicate}) for each object. The multiplication of two scores $s_0 s_1$ is the final classification score. Therefore, a good detection should have both scores large.

The network has three steps. First, the $1024$-d feature and classification score is fused to generate the appearance feature. Second, a relation module transforms such appearance features of all objects. Last, the transformed features of each object pass a linear classifier ($W_{s}$ in Figure~\ref{fig.duplicate_detection_with_relation} (b)) and sigmoid to output the probability $\in[0, 1]$.

The relation module is at the core of the network. It enables effective end-to-end learning using information from multiple sources (the bounding boxes, original appearance features and classification scores). In addition, the usage of the classification scores also turns out important.

\textbf{Rank feature} We found that it is most effective to transform the score into a rank, instead of using its value. Specifically, the input $N$ objects are sorted in descending order of their scores. Each object is given a rank $\in[1,N]$ accordingly. The scalar rank is then embedded into a higher dimensional $128$-d feature, using the same method~\cite{vaswani2017attention} as for geometry feature embedding in Section~\ref{sec.relation_module}.

Both the rank feature and original $1024$-d appearance feature are transformed to $128$-d (via $W_{f_R}$ and $W_f$ in Figure~\ref{fig.duplicate_detection_with_relation} (b), respectively), and added as the input to the relation module.

\textbf{Which object is correct?} Given a number of detected objects, it is not immediately clear which one should be matched to a ground truth object as \emph{correct}. The most obvious choice would be following the evaluation criterion of Pascal VOC~\cite{everingham2010pascal} or COCO datasets~\cite{lin2014coco}. That is, given a predefined threshold $\eta$ for the IoU between detection box and ground truth box,  all detection boxes with $\text{IoU} \geq \eta$  are firstly matched to the same ground truth. The detection box with highest score is \emph{correct} and others are \emph{duplicate}.

Consequently, such selection criteria work best when learning and evaluation use the same threshold $\eta$  . For example, using $\eta=0.5$ in learning produces best mAP$@0.5$ metric but not mAP $@0.75$. This is verified in Table~\ref{table.exp_duplicate_removal_nms_comparison}.

This observation suggests a unique benefit of our approach that is missing in NMS: the duplicate removal step can be adaptively learnt according to needs, instead of using preset parameters. For example, a large $\eta$ should be used when a high localization accuracy is desired.

Motivated by the COCO evaluation criteria (mAP$@0.5-0.95$),  our best practice is to use multiple thresholds simultaneously, i.e., $\eta\in\{0.5,0.6,0.7,0.8,0.9\}$. Specifically, the classifier $W_s$ in Figure.~\ref{fig.duplicate_detection_with_relation} (b) is changed to output multiple probabilities corresponding to different IoU thresholds and \emph{correct} detections, resulting in multiple binary classification loss terms. The training is well balanced between different cases. During inference, the multiple probabilities are simply averaged as a single output.

\textbf{Training} The binary cross entropy loss is used on the final score (multiplication of two scores, see Figure~\ref{fig.duplicate_detection_with_relation} (b)). The loss is averaged over all detection boxes on all object categories. A single network is trained for all object categories.

Note that the duplicate classification problem is extremely imbalanced. Most detections are \textit{duplicate}. The ratio of \textit{correct} detections is usually $<0.01$. Nevertheless, we found the simple cross entropy loss works well. This is attributed to the multiplicative behavior in the final score $s_0 s_1$. Because most detections have very small $s_0$ (mostly $<0.01$) and thus small $s_0 s_1$. The magnitude of their loss values $L=-\log(1-s_0s_1)$ (for non-\textit{correct} object) and back-propagated gradients $\partial{L}/\partial{s_1}=s_0/(1-s_0 s_1)$ is also very small and does not affect the optimization much. Intuitively, training is focused on a few real \textit{duplicate} detections with large $s_0$. This shares the similar spirit to the recent focal loss work~\cite{DBLP:journals/corr/abs-1708-02002}, where majority insignificant loss terms are down weighted and play minor roles during optimization.

\textbf{Inference} The same duplicate removal network is applied for all object categories independently. At a first glance, the runtime complexity could be high, when the number of object classes (80 for COCO dataset~\cite{lin2014coco}) and detections ($N=300$) is high. Nevertheless, in practice most detections' original score $s_0$ is nearly $0$ in most object classes. For example, in the experiments in Table~\ref{table.exp_duplicate_removal_nms_comparison}, only $12.0\%$ classes have detection scores $>0.01$ and in these classes only $6.8\%$ detections have scores $>0.01$.

After removing these insignificant classes and detections, the final recognition accuracy is not affected. Running the duplicate removal network on remaining detections is practical, taking about $2$ ms on a Titan X GPU. Note that NMS and SoftNMS~\cite{bodla2017soft} methods are sequential and take about $5$ ms on a CPU~\cite{bodla2017soft}. Also note that the recent learning NMS work~\cite{hosang2017learning} uses a very deep and complex network (depth up to 80), which is much less efficient than ours.

\subsection{End-to-End Object Detection}
\label{sec.end_to_end_detection}
\vspace{-.5em}

The duplicate removal network is trained alone in Section~\ref{sec.duplicate_removal}. Nevertheless, there is nothing preventing the training to be end-to-end. As indicated by the red arrows in Figure~\ref{fig.duplicate_detection_with_relation} (b), the back propagated gradients can pass into the original $1024$-d features and classification scores, which can further propagate back into the \emph{head} and \emph{backbone} networks.

Our end-to-end training simply combines the region proposal loss, the instance recognition loss in Section~\ref{sec.feature_enhance} and duplicate classification loss in Section~\ref{sec.duplicate_removal}, with equal weights. For instance recognition, either the original head Eq.~(\ref{eq.instance_recognition_2fc_head}) or enhanced head Eq.~(\ref{eq.instance_recognition_relation}) can be used.

The end-to-end training is clearly feasible, but does it work? At a first glance, there are two issues.

First, the goals of instance recognition step and duplicate removal step seem contradictory. The former expects all objects matched to the same ground truth object to have high scores. The latter expects only one of them does. In our experiment, we found the end-to-end training works well and converges equally fast for both networks, compared to when they are trained individually as in Section~\ref{sec.feature_enhance} and~\ref{sec.duplicate_removal}. We believe this seemingly conflict is reconciled, again, via the multiplicative behavior in the final score $s_0 s_1$, which makes the two goals complementary other than conflicting. The instance recognition step only needs to produce high score $s_0$ for good detections (no matter duplicate or not). The duplicate removal step only needs to produce low score $s_1$ for duplicates. The majority non-object or \textit{duplicate} detection is correct as long as one of the two scores is correct.

Second, the binary classification ground truth label in the duplicate removal step depends on the output from the instance recognition step, and changes during the course of end-to-end training. However, in experiments we did not observe adverse effects caused by this instability. While there is no theoretical evidence yet, our guess is that the duplicate removal network is relatively easy to train and the instable label may serve as a means of regularization.

As verified in experiments (Section~\ref{sec.exp_end_to_end_detection}), the end-to-end training improves the recognition accuracy.

\section{Experiments}
\vspace{-.5em}

All experiments are performed on COCO detection datasets with 80 object categories~\cite{lin2014coco}. A union of $80k$ train images and a $35k$ subset of val images are used for training~\cite{bell2016inside,lin2016feature}. Most ablation experiments report detection accuracies on a subset of $5k$ unused val images (denoted as \emph{minival}) as common practice~\cite{bell2016inside,lin2016feature}. Table~\ref{table.exp_system_improvement} also reports accuracies on $test$-$dev$ for system-level comparison.

For backbone networks, we use ResNet-50 and ResNet-101~\cite{he2016deep}. Unless otherwise noted, ResNet-50 is used.

For Faster RCNN~\cite{ren2015faster} and DCN~\cite{dai2017deformable}, our training mostly follow~\cite{dai2017deformable}. For FPN~\cite{lin2016feature}, our training mostly follow~\cite{lin2016feature}. See Appendix for details.

\subsection{Relation for Instance Recognition}
\label{sec.exp_feature_enhance}
\vspace{-.5em}

\setlength{\tabcolsep}{3pt}
\renewcommand{\arraystretch}{1.1}
\begin{table*}[t]
\small
\begin{center}
\begin{tabular}{c|ccc|cccccc|ccccc}
\hline
\tabincell{c}{2fc baseline}& \multicolumn{3}{c|}{\emph{\footnotesize \textbf{(a)}: usage of geometric feature}} & \multicolumn{6}{c|}{\emph{\footnotesize \textbf{(b)}: number of relations $N_r$}} & \multicolumn{5}{c}{\emph{\footnotesize \textbf{(c)}: number of relation modules $\{r_1,r_2\}$}} \\
\hline
 & none & unary & ours* & 1 & 2 & 4 & 8 & 16* & 32 & {$\{1,0\}$} & {$\{0,1\}$} & {$\{1,1\}$*} & {$\{2,2\}$} & {$\{4,4\}$} \\
\hline
29.6 & 30.3 & 31.1 & \textbf{31.9} & 30.5 & 30.6 & 31.3 & 31.7 & \textbf{31.9} & 31.7 & 31.7 & 31.4 & 31.9 & 32.5 & \textbf{32.8} \\
\hline
\end{tabular}
\end{center}
\vspace{-.5em}
\caption{Ablation study of relation module structure and parameters (* for default). mAP@all is reported.}
\label{table.exp_feature_enhance_module_structure}
\vspace{-1em}
\end{table*}

\setlength{\tabcolsep}{3pt}
\renewcommand{\arraystretch}{1.1}
\begin{table}[t]
\small
\begin{center}
\begin{tabular}{l|ccc|cc}
\tabincell{c}{head} & \scriptsize \tabincell{c}{mAP} &
\scriptsize mAP$_{50}$ & \scriptsize mAP$_{75}$ & \scriptsize \# params & \scriptsize \# FLOPS \\
\hline
(a) 2fc (1024) & 29.6 & 50.9 & 30.1 & 38.0M & 80.2B \\
\hline
(b) 2fc (1432) & 29.7 & 50.3 & 30.2 & 44.1M & 82.0B \\
(c) 3fc (1024) & 29.0 & 49.4 & 29.6 & 39.0M & 80.5B \\
(d) {\scriptsize 2fc+res $\{r_1,r_2\}$=$\{1,1\}$}  & 29.9 & 50.6 & 30.5 & 44.0M & 82.1B \\
(e) 2fc (1024) + global & 29.6 & 50.3 & 30.8 & 38.2M & 82.2B \\
(f) {\scriptsize 2fc+RM $\{r_1,r_2\}$=$\{1,1\}$} & \textbf{31.9} & 53.7 & 33.1 & 44.0M & 82.6B  \\
\hline
(g) 2fc (1024) + $2\times$ & 30.4 & 51.7 & 31.4 & 50.2M & 83.8B \\
(h) {\scriptsize 2fc+$2\times$+RM $\{r_1,r_2\}$=$\{1,1\}$} & \textbf{32.5} & 54.3 & 34.1 & 56.2M & 86.2B \\
\hline
(i) {\scriptsize 2fc+res $\{r_1,r_2\}$=$\{2,2\}$}  & 29.8 & 50.5 & 30.5 & 50.0M & 84.0B\\  % need update
(j) {\scriptsize 2fc+RM $\{r_1,r_2\}$=$\{2,2\}$}  & \textbf{32.5} & 54.0 & 33.8 & 50.0M & 84.9B \\
\hline
\end{tabular}
\end{center}
\caption{Comparison of various heads with similar complexity.}
\label{table.exp_feature_enhance_baseline}
\vspace{-.5em}
\end{table}

In this section, NMS with IoU threshold of 0.6 is used for duplicate removal for all experiments.

\textbf{Relation module improves instance recognition} Table~\ref{table.exp_feature_enhance_module_structure} compares the baseline $2fc$ head in Eq.~(\ref{eq.instance_recognition_2fc_head})  with the proposed $2fc+RM$ head in Eq.~(\ref{eq.instance_recognition_relation}), under various parameters.

We firstly note that our baseline implementation achieves reasonable accuracy ($29.6$ mAP) when compared with the literature (\eg, \cite{lin2016feature} reports $28.0$ using ResNet-50 and \cite{dai2017deformable} reports $29.4$ using ResNet-101).

Ablation studies are performed on three key parameters.

\emph{Usage of geometric feature.} As analyzed in Section~\ref{sec.relation_module}, our usage of geometric feature in Eq.~(\ref{eq.geometric_weight}) is compared to two plain implementations. Results show that our approach is the best, although all the three surpass the baseline.

\emph{Number of relations $N_r$.} Using more relations steadily improves the accuracy. The improvement saturates at $N_r=16$, where +2.3 mAP gain is achieved.

\emph{Number of modules.} Using more relation modules steadily improves accuracy, up to +3.2 mAP gain. As this also increases the parameter and computation complexity, by default $r_1=1, r_2=1$ is used.

\setlength{\tabcolsep}{3pt}
\renewcommand{\arraystretch}{1.1}
\begin{table}[t]
\small
\begin{center}
\begin{tabular}{c|c|cc|c|cc}
\hline
\tabincell{c}{\footnotesize NMS}& \tabincell{c}{\footnotesize ours} & \multicolumn{2}{c|}{\footnotesize rank $f_R$} & \multicolumn{1}{c|}{\footnotesize appearance $f$} & \multicolumn{2}{c}{\footnotesize geometric $bbox$} \\
\hline
& $\{f_R,f,bbox\}$ & \textit{none} & $s_0$ & \textit{none} & \textit{none} & \textit{unary} \\
\hline
29.6 & \textbf{30.3} & 26.6 & 28.3 & 29.9 & 28.1 & 28.2 \\
\hline
\end{tabular}
\end{center}
\vspace{-.5em}
\caption{Ablation study of input features for duplicate removal network (\emph{none} indicates without such feature).}
\label{table.exp_duplicate_removal_features}
\vspace{-1em}
\end{table}

\setlength{\tabcolsep}{3pt}
\renewcommand{\arraystretch}{1.1}
\begin{table}[t]
\small
\begin{center}
\begin{tabular}{c|c|ccc}
\tabincell{c}{method}&\tabincell{c}{parameters} & \tabincell{c}{mAP} &
mAP$_{50}$ & mAP$_{75}$ \\
\hline
NMS & $N_t=0.3$ & 29.0 &51.4    & 29.4   \\
NMS & $N_t=0.4$ & 29.4 & \textbf{52.1}  & 29.5   \\
NMS & $N_t=0.5$ & 29.6 & 51.9   & 29.7   \\
NMS & $N_t=0.6$ & \textbf{29.6} & 50.9  & 30.1   \\
NMS & $N_t=0.7$ & 28.4 & 46.6   & \textbf{30.7}   \\
\hline
SoftNMS & $\sigma=0.2$ & 30.0   &  \textbf{52.3}        & 30.5  \\
SoftNMS & $\sigma=0.4$ & 30.2   &  51.7 & 31.3  \\
SoftNMS & $\sigma=0.6$ &\textbf{30.2}   &  50.9 & 31.6  \\
SoftNMS & $\sigma=0.8$ & 29.9   &  49.9 & \textbf{31.6}  \\
SoftNMS & $\sigma=1.0$ & 29.7   &  49.7 & 31.6  \\
\hline
\tabincell{c}{ours} & $\eta=0.5$ & 30.3 &  \textbf{51.9}        & 31.5  \\
\tabincell{c}{ours}  & $\eta=0.75$ & 30.1 &  49.0 & \textbf{32.7}  \\
\tabincell{c}{ours} &$\eta\in[0.5, 0.9]$ & \textbf{30.5} & 50.2 & 32.4  \\
\hline
\tabincell{c}{ours (e2e)} &{$\eta \in[0.5, 0.9]$} & \textbf{31.0} & 51.4 & 32.8  \\
\hline
\end{tabular}
\end{center}
\caption{Comparison of NMS methods and our approach (Section~\ref{sec.duplicate_removal}). Last row uses end-to-end training (Section~\ref{sec.end_to_end_detection}).}
\vspace{-1em}
\label{table.exp_duplicate_removal_nms_comparison}
\vspace{-.5em}
\end{table}
\textbf{Does the improvement come from more parameters or depths?} Table~\ref{table.exp_feature_enhance_baseline} answers this question by enhancing the baseline $2fc$ head (a) in width or depth such that its complexity is comparable to that of adding relation modules.

A wider 2fc head ($1432$-d, b) only introduces small improvement (+0.1 mAP). A deeper 3fc head (c) deteriorates the accuracy (-0.6 mAP), probably due to the difficulty of training. To make the training easier, residual blocks~\cite{he2016deep} are used\footnote{Each residual branch in a block has three 1024-d fc layers to have similar complexity as an object relation module. The residual blocks are inserted at the same positions as our object relation modules.} (d), but only moderate improvement is observed (+0.3 mAP). When global context is used (e, $2048$-d global average pooled features are concatenated with the second $1024$-d instance feature before classification), no improvement is observed. By contrast, our approach (f) significantly improves the accuracy (+2.3 mAP).

We also consider another baseline which concatenates the original pooled features with the ones from a $2\times$ larger RoI (g), the performance is improved from 29.6 to 30.4 mAP, indicating a better way of utilizing context cues. In addition, we combine this new head with relation modules, that is, replacing the \textit{2fc} with $\{r_1,r_2\}$=$\{1,1\}$ (h). We get 32.5 mAP, which is 0.6 better than setting (f) (31.9 mAP). This indicates that using a larger window context and relation modules are mostly complementary.

When more residual blocks are used and the head network becomes deeper (i), accuracy no longer increases. While, accuracy is continually improved when more relation modules are used (j).

The comparison indicates that the relation module is effective and the effect is beyond increasing network capacity.

\setlength{\tabcolsep}{2pt}
\renewcommand{\arraystretch}{1.1}
\begin{table*}[t]
\small
\begin{center}
\begin{tabular}{c|c|ccc|cc}
\tabincell{c}{backbone} &  test set & \tabincell{c}{ mAP} &
{mAP$_{50}$} & {mAP$_{75}$} & {\#. params} & {FLOPS} \\
\shline
\multirow{2}{*}{faster RCNN~\cite{ren2015faster}} & \emph{minival} & 32.2$\rightarrow$34.7$\rightarrow$\textbf{35.2} &52.9$\rightarrow$55.3$\rightarrow$\textbf{55.8}        & 34.2$\rightarrow$37.2$\rightarrow$\textbf{38.2} & \multirow{2}{*}{58.3M$\rightarrow$64.3M$\rightarrow$64.6M} & \multirow{2}{*}{122.2B$\rightarrow$124.6B$\rightarrow$124.9B} \\
& \emph{test-dev} & 32.7$\rightarrow$35.2$\rightarrow$\textbf{35.4} &53.6$\rightarrow$\textbf{56.2}$\rightarrow$56.1    & 34.7$\rightarrow$37.8$\rightarrow$\textbf{38.5} &  &\\
\hline
\multirow{2}{*}{FPN~\cite{lin2016feature}} & \emph{minival} & 36.8$\rightarrow$38.1$\rightarrow$\textbf{38.8} & 57.8$\rightarrow$59.5$\rightarrow$\textbf{60.3}       & 40.7$\rightarrow$41.8$\rightarrow$\textbf{42.9}& \multirow{2}{*}{56.4M$\rightarrow$62.4M$\rightarrow$62.8M} & \multirow{2}{*}{145.8B$\rightarrow$157.8B$\rightarrow$158.2B}\\
 & \emph{test-dev} & 37.2$\rightarrow$38.3$\rightarrow$\textbf{38.9} & 58.2$\rightarrow$59.9$\rightarrow$\textbf{60.5} & 41.4$\rightarrow$42.3$\rightarrow$\textbf{43.3} & & \\
\hline
\multirow{2}{*}{DCN~\cite{dai2017deformable}} & \emph{minival} & 37.5$\rightarrow$38.1$\rightarrow$\textbf{38.5} & 57.3$\rightarrow$57.8$\rightarrow$\textbf{57.8}       & 41.0$\rightarrow$41.3$\rightarrow$\textbf{42.0} & \multirow{2}{*}{60.5M$\rightarrow$66.5M$\rightarrow$66.8M} & \multirow{2}{*}{125.0B$\rightarrow$127.4B$\rightarrow$127.7B} \\
 & \emph{test-dev} & 38.1$\rightarrow$38.8$\rightarrow$\textbf{39.0} & 58.1$\rightarrow$\textbf{58.7}$\rightarrow$58.6 & 41.6$\rightarrow$42.4$\rightarrow$\textbf{42.9}  &  &  \\
\end{tabular}
\end{center}
\vspace{-.5em}
\caption{Improvement (2fc head+SoftNMS~\cite{bodla2017soft}, 2fc+RM head+SoftNMS and 2fc+RM head+e2e from left to right connected by $\rightarrow$) in state-of-the-art systems on COCO \emph{minival} and \emph{test-dev}. Online hard example mining (OHEM)~\cite{shrivastava2016training} is adopted. Also note that the strong SoftNMS method ($\sigma=0.6$) is used for duplicate removal in non-e2e approaches.}
\vspace{-1em}
\label{table.exp_system_improvement}
\end{table*}

\textbf{Complexity} In each relation module, $d_f=1024, d_k=64, d_g=64$. When $N_r=16$, a module has about 3 million parameters and 1.2 billion FLOPs, as from Eq.~(\ref{eq.space_complexity}) and (\ref{eq.compute_complexity}). The computation overhead is relatively small, compared to the complexity of whole detection networks as shown in Table.~\ref{table.exp_system_improvement} (less than 2\% for faster RCNN~\cite{ren2015faster} / DCN~\cite{dai2017deformable} and about 8\% for FPN~\cite{lin2016feature}).

\subsection{Relation for Duplicate Removal}
\label{sec.exp_duplicate_removal}
\vspace{-.5em}

All the experiments in this section use the detected objects of the Faster RCNN baseline $2fc$ head in Table~\ref{table.exp_feature_enhance_module_structure} (top row, $29.6$ mAP after NMS) for training and inference of our approach in Section~\ref{sec.duplicate_removal}.

In our approach, the relation module parameters are set as $d_f=128, d_k=64, d_g=64, N_r=16, N=100$. Using larger values no longer increases accuracy. The duplicate removal network has 0.33 million parameters and about 0.3 billion FLOPs. This overhead is small, about 1\% in both model size and computation compared to a faster RCNN baseline network with ResNet-50.

Table~\ref{table.exp_duplicate_removal_features} investigates the effects of different input features to the relation module (Figure~\ref{fig.duplicate_detection_with_relation} (b)). Using $\eta=0.5$, our approach improves the mAP to $30.3$. When the rank feature is not used, mAP drops to $26.6$. When the class score $s_0$ replaces the rank in a similar way (the score is embedded to $128$-d), mAP drops to $28.3$. When $1024$-d appearance feature is not used, mAP slightly drops to $29.9$. These results suggest that rank feature is most crucial for final accuracy.

When geometric feature is not used, mAP drops to $28.1$. When it is used by \textit{unary} method as mentioned in Section~\ref{sec.relation_module} and in Table~\ref{table.exp_feature_enhance_module_structure} (a), mAP drops to $28.2$. These results verify the effectiveness of our usage of geometric weight Eq.~(\ref{eq.geometric_weight}).

\textbf{Comparison to NMS}
Table~\ref{table.exp_duplicate_removal_nms_comparison} compares our method with NMS method and its better variant SoftNMS~\cite{bodla2017soft}, which is also the state-of-the-art method for duplicate removal.

Note that all three methods have a single parameter of similar role of controlling the localization accuracy: the IoU threshold $N_t$ in NMS, the normalizing parameter $\sigma$ in SoftNMS~\cite{bodla2017soft}, and the ground truth label criteria parameter $\eta$ in ours. Varying these parameters changes accuracy under different localization metrics. However, it is unclear how to set the optimal parameters for NMS methods, other than trial-and-error. Our approach is easy to interpret because the parameter $\eta$ directly specify the requirement on localization accuracy. It performs best for mAP$_{50}$ when $\eta=0.5$, for mAP$_{75}$ when $\eta=0.75$, and for mAP when $\eta\in[0.5,0.9]$.

Our final mAP accuracy is better than NMS and SoftNMS, establishing the new state-of-the-art. In the following end-to-end experiments, $\eta\in[0.5,0.9]$ is used.

\setlength{\tabcolsep}{0.5pt}
\begin{figure}
\begin{center}
\begin{tabular}{c}
{\includegraphics[width=0.48\textwidth]{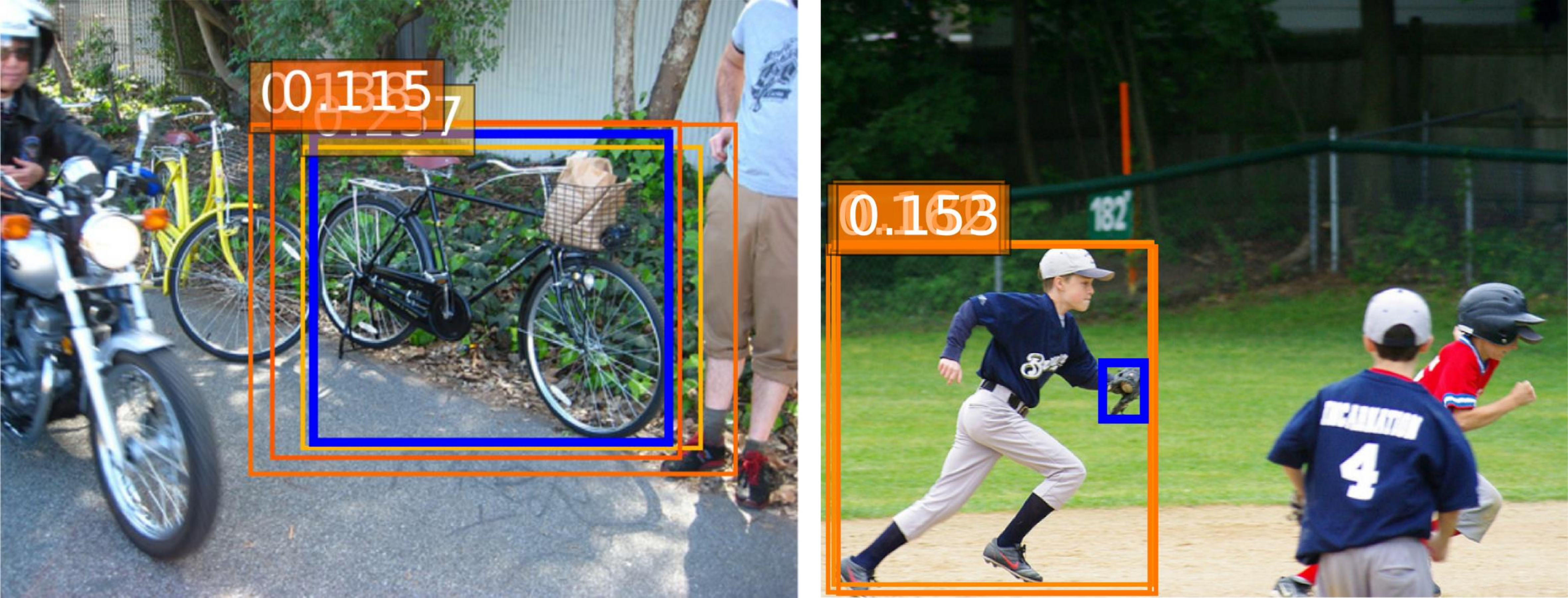}}  \\
\end{tabular}
\end{center}
\vspace{-.5em}
\caption{Representative examples with high relation weights in Eq.~(\ref{eq.object_relation_weight}). The reference object $n$ is blue. The other objects contributing a high weight (shown on the top-left) are yellow.}
\label{figure.relation_examples_v2}
\vspace{-1em}
\end{figure}

\subsection{End-to-End Object Detection}
\label{sec.exp_end_to_end_detection}
\vspace{-.5em}

The last row in Table~\ref{table.exp_duplicate_removal_nms_comparison} compares the end-to-end learning with separate training of instance recognition and duplicate removal. The end-to-end learning improves the accuracy by +0.5 mAP.

Finally, we investigate our approach on some stronger backbone networks, i.e., ResNet-101~\cite{he2016deep} and better detection architectures, i.e., FPN~\cite{lin2016feature} and DCN~\cite{dai2017deformable} in Table~\ref{table.exp_system_improvement}. Using faster RCNN with ResNet-101, by replacing the 2fc head with 2fc+RM head in Table~\ref{table.exp_feature_enhance_module_structure} (default parameters), our approach improves by 2.5 mAP on COCO \textit{minival}. Further using duplicate removal network with end2end training, the accuracy improves further by 0.5 mAP. The improvement on COCO \emph{test-dev} is similar. On stronger baselines, e.g., DCN~\cite{dai2017deformable} and FPN~\cite{lin2016feature}, we also have moderate improvements on accuracy by both feature enhanced network and duplicate removal with end2end training. Also note that our implementation of baseline networks has higher accuracy than that in original works (38.1 versus 33.1~\cite{dai2017deformable}, 37.2 versus 36.2~\cite{lin2016feature}).

\vspace{-.5em}
\section{Conclusions}
\vspace{-.5em}

The comprehensive ablation experiments suggest that the relation modules have learnt information between objects that is missing when learning is performed on individual objects. Nevertheless, it is not clear what is learnt in the relation module, especially when multiple ones are stacked.

Towards understanding, we investigate the (only) relation module in the $\{r_1,r_2\}=\{1,0\}$ head in Table~\ref{table.exp_feature_enhance_module_structure}(c). Figure~\ref{figure.relation_examples_v2} show some representative examples with high relation weights. The left example suggests that several objects overlapping on the same ground truth (bicycle) contribute to the centering object. The right example suggests that the person contributes to the glove. While these examples are intuitive, our understanding of how relation module works is preliminary and left as future work.

\appendix

\renewcommand{\thesection}{A\arabic{section}}

\section{Training Details}

For Faster RCNN~\cite{ren2015faster} and DCN~\cite{dai2017deformable}, the hyper-parameters in training mostly follow~\cite{dai2017deformable}. Images are resized such that their shorter side is $600$ pixels. The number of region proposals $N$ is 300. 4 scales and 3 aspect ratios are adopted for anchors. Region proposal and instance recognition networks are jointly trained. Both instance recognition (Section~\ref{sec.exp_feature_enhance}) and end-to-end (Section~\ref{sec.exp_end_to_end_detection}) training have $\sim450k$ iterations (8 epochs). Duplicate removal (Section~\ref{sec.exp_duplicate_removal}) training has $\sim170k$ iterations (3 epochs). The learning rates are set as $2\times 10^{-3}$ for the first $\frac{2}{3}$ iterations and $2\times 10^{-4}$ for the last $\frac{1}{3}$ iterations.

For FPN~\cite{lin2016feature}, hyper-parameters in training mostly follow~\cite{lin2016feature}. Images are resized such that their shorter side is $800$ pixels. The number of region proposals $N$ is $1000$\footnote{In~\cite{lin2016feature}, 2000 are used for training while 1000 are used for test. Here we use 1000 in both training and test for consistency.}. 5 scales and 3 aspect ratios are adopted for anchors. Region proposal network is trained for about $170k$ iterations ($3$ epochs). Both instance recognition (Section~\ref{sec.exp_feature_enhance}) and end-to-end training (Section~\ref{sec.exp_end_to_end_detection}) have $\sim 340k$ iterations (6 epochs). The learning rates are set as $5\times 10^{-3}$ for the first $\frac{2}{3}$ iterations and $5 \times 10^{-4}$ for the last $\frac{1}{3}$ iterations.

For all training, SGD is performed on 4 GPUs with 1 image on each. Weight decay is $1 \times 10^{-4}$ and momentum is $0.9$. Class agnostic bounding box regression~\cite{dai2016rfcn} is adopted as it has comparable accuracy with the class aware version but higher efficiency.

For \emph{instance recognition} subnetwork, all $N$ proposals are used to compute loss. We find it has similar accuracy with the usual practice that a subset of sampled proposals are used~\cite{girshick2015fast,ren2015faster,dai2016rfcn,lin2016feature,dai2017deformable} (In~\cite{dai2017deformable}, 128 are sampled from 300 proposals and positive negative ratio is coarsely guaranteed to be 1:3. 512 are sampled from 2000 proposals in~\cite{lin2016feature}). We also consider online hard example mining (OHEM)~\cite{shrivastava2016training} approach in Table ~\ref{table.exp_system_improvement} for better overall baseline performance. For Faster RCNN and DCN, 128 hard examples are sampled from 300 proposals. For FPN, 512 are sampled from 1000 proposals.

{\small
\bibliographystyle{ieee}
\bibliography{egbib}
}

\end{document}